\documentclass{article}
\usepackage{multirow}
\usepackage{booktabs}
\usepackage{graphicx}
\usepackage{subfigure}
\usepackage{comment}
\usepackage{afterpage}
\usepackage[bottom]{footmisc}
\usepackage[driverfallback=dvipdfm,  
pdfstartview=FitH,
CJKbookmarks=true,
bookmarksnumbered=true,
bookmarksopen=true,
colorlinks, 
pdfborder=001,   
linkcolor=blue,
anchorcolor=green,
citecolor=green
]{hyperref}
\usepackage{spconf,amsmath,graphicx}

\let\oldmaketitle\maketitle
\renewcommand{\maketitle}{\oldmaketitle}
\title{HANDLING NOISE IN IMAGE DEBLURRING VIA JOINT LEARNING}
\name{Si Miao, Yongxin Zhu}
\address{Shanghai Advanced Research Institute, Chinese Academy of Sciences, China}
\begin{document}
\topmargin=0mm
%
\maketitle
\setcounter{footnote}{0}
\begin{abstract}
Currently, many blind deblurring methods assume blurred images are noise-free and perform unsatisfactorily on the blurry images with noise. Unfortunately, noise is quite common in real scenes. A straightforward solution is to denoise images before deblurring them. However, even state-of-the-art denoisers cannot guarantee to remove noise entirely. Slight residual noise in the denoised images could cause significant artifacts in the deblurring stage. To tackle this problem, we propose a cascaded framework consisting of a denoiser subnetwork and a deblurring subnetwork. In contrast to previous methods, we train the two subnetworks jointly. Joint learning reduces the effect of the residual noise after denoising on deblurring, hence improves the robustness of deblurring to heavy noise. Moreover, our method is also helpful for blur kernel estimation. Experiments on the CelebA dataset and the GOPRO dataset show that our method performs favorably against several state-of-the-art methods. 
\end{abstract}
\section{Introduction}
\label{sec:intro}
This work is on blind deblurring of a single blurry image with noise. The fundamental blur model is:
{
	\setlength\abovedisplayskip{1pt plus 3pt minus 7pt}
	\setlength\belowdisplayskip{1pt plus 3pt minus 7pt}
	\begin{equation}
	N = I*P+n \label{1},
	\end{equation}
}
where $N$ is the blurred image, $I$ is the sharp image, $*$ is the convolution operator, $P$ is the blur kernel, and $n$ is the noise term. The blur kernel $P$ is also known as the point spread function (PSF). Priors based approaches and deep learning based approaches are two major kinds of approaches to blind deblurring.

Priors based approaches, e.g., \cite{DC_Deblur, classPrior}, are usually based on the uniform blur model (Eq. \ref{1}) that assumes the blur kernels $P$ are spatial-invariant. However, most motion blurs in real scenes are non-uniform because different objects have diverse moving trajectories. Deep models like DeblurGAN \cite{DeblurGAN}, SRN \cite{SRN}, GFN \cite{GFN}, and Inception GAN \cite{Inception_GAN} are excellent at deblurring noise-free images with complex non-uniform blurs. Nevertheless, they are trained on noisy-free images and could hardly deblur noisy images (see Fig. \ref{1}(b)). A straightforward idea is to denoise these noisy images before deblurring them. The idea has two major problems. First, it is common that denoised images still contain slight noise (Fig. \ref{1}(e)). The slight noise are propagated into the deblurring networks and jeopardize the deblurring stage (Fig. \ref{1}(f)). Second, denoisers (e.g., BM3D \cite{bm3d} and DnCNN \cite{DnCNN}) usually rely on noise level estimation that would lead to significant artifacts if the estimation is inaccurate. If we underestimate the noise level, the denoised image would remain noisy (Fig. \ref{1}(c, d)). If we overestimate the noise level, the denoised image would be oversmoothed and blurrier.

To our knowledge, there are few prior works to handle noise in image deblurring. One important attempt \cite{directional} combined directional filtering with the noise-aware kernel estimation algorithm. However, their work was limited to uniform blurs and slight noise. Anger et al. \cite{l0_noise} proposed refining the $\ell_0$ prior \cite{PanL0}. Despite strong robustness and short running time, their work was also limited to uniform blurs. In this work, we propose a \textbf{N}oisy \textbf{I}mages \textbf{D}eblurring \textbf{F}ramework (NIDF) composed of a denoiser subnetwork and a deblurring subnetwork cascaded in series. Specifically, we propose a loss function $\mathcal{L}_{joint}$ (Eq. \ref{joint loss}) to train the two subnetworks jointly. Better than most deblurring methods that are not noise-robust, NIDF could generate sharp images from blur images with the presence of noise. Different from DnCNN \cite{DnCNN} that trains a corresponding model for each noise level, we train NIDF under mixed noise levels. As a result, NIDF adapts to various noise intensities and does not require noise level estimation during training and inference. Extensive experiments on the CelebA \cite{CelebA} dataset and the GOPRO \cite{DeepDeblur} dataset show that joint learning significantly improves performance. 

\footnotetext[1]{Code, data and more experimental results available at: \url{https://github.com/miaosiSari/face_and_natural_images_deblurring}\label{link}}

\begin{figure*}[!t]
	\centering
	\begin{tabular}{cccc}
		\centering
		\subfigure[Input (noise level $\sigma$=30)]{
			\begin{minipage}[!t]{0.2\linewidth}
				\centering
				\includegraphics[width=1.2in,height=0.6in]{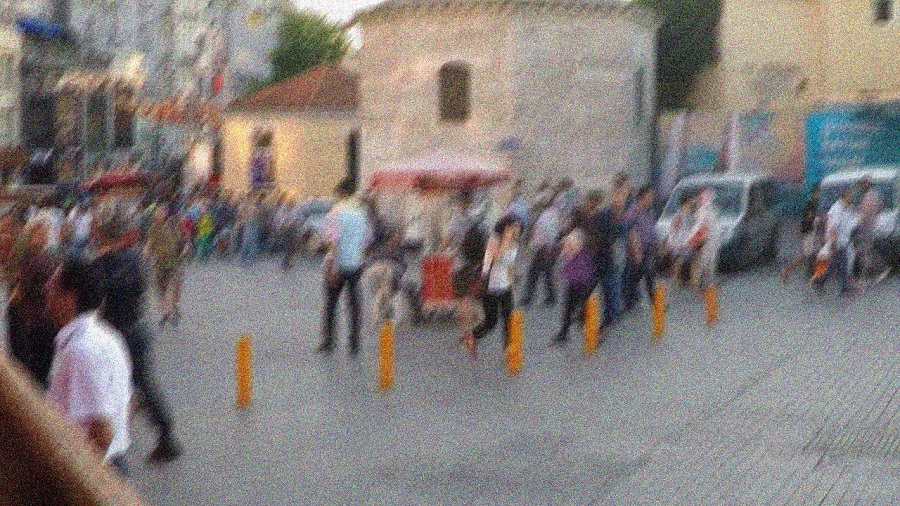}
			\end{minipage}%
		}&%
		\subfigure[SRN]{
			\begin{minipage}[!t]{0.2\linewidth}
				\centering
				\includegraphics[width=1.2in,height=0.6in]{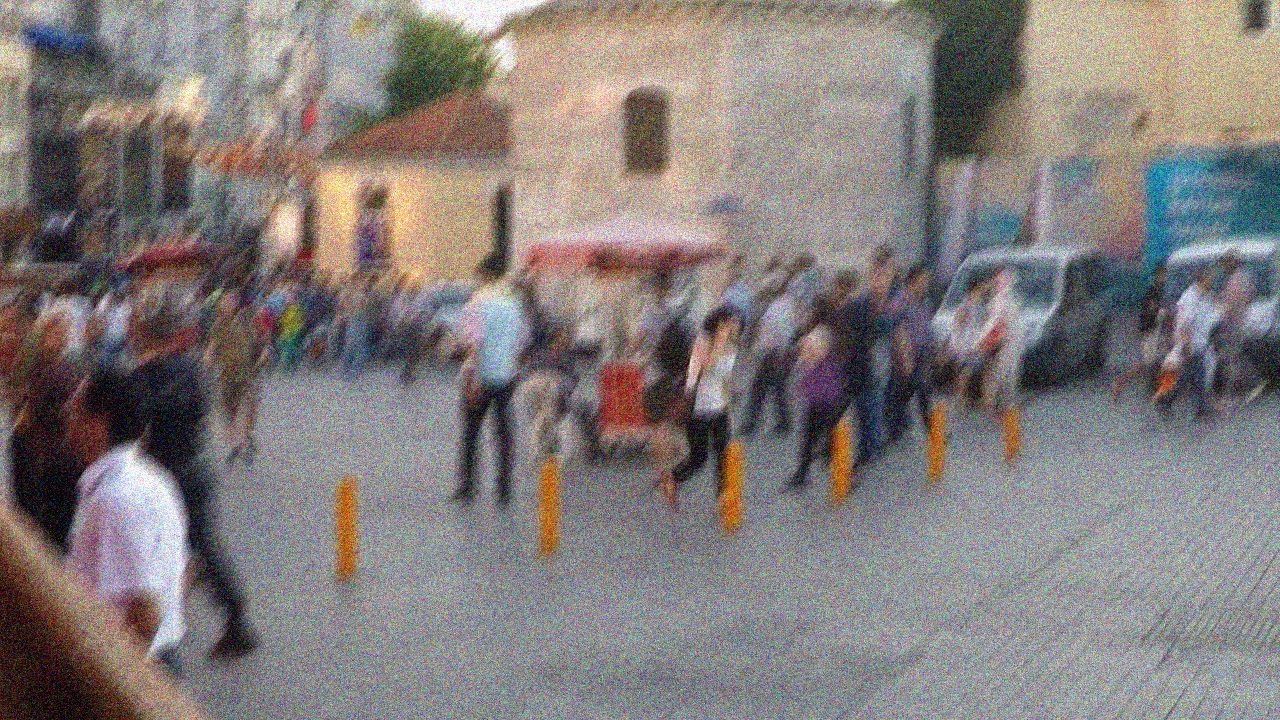}
			\end{minipage}
		}&%
		\subfigure[BM3D ($\sigma=20$)]{
			\begin{minipage}[!t]{0.2\linewidth}
				\centering
				\includegraphics[width=1.2in,height=0.6in]{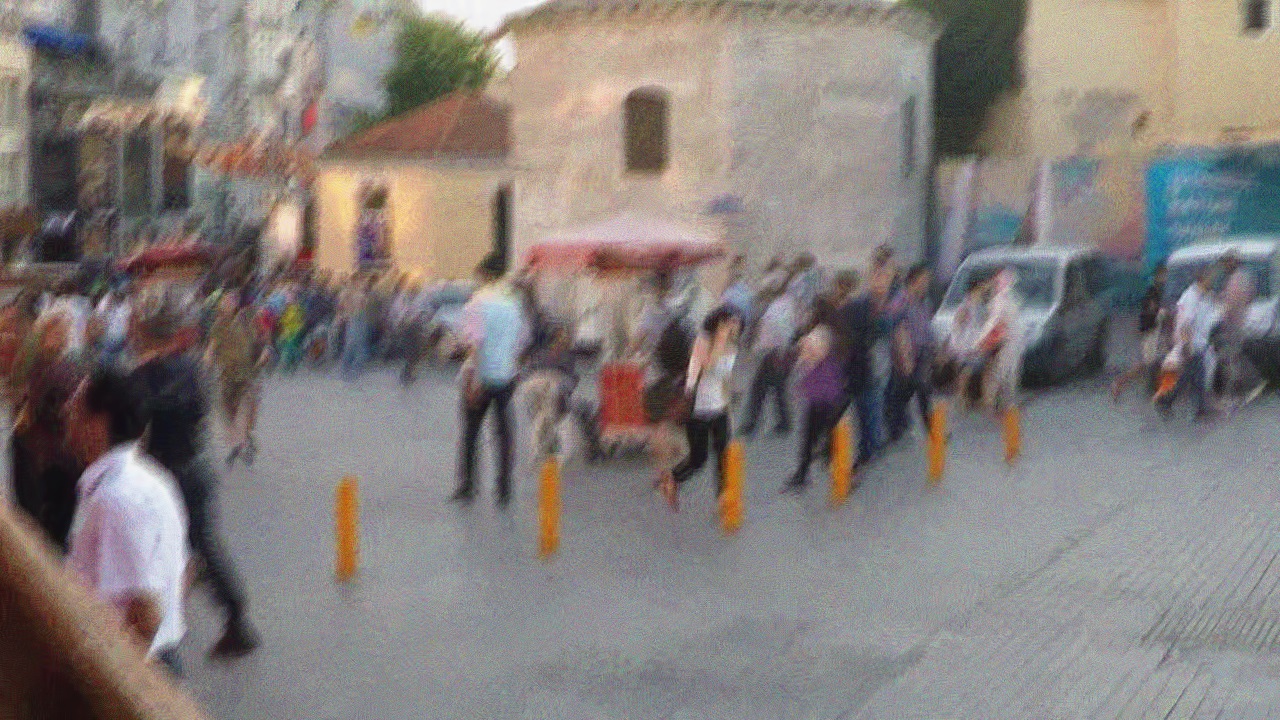}
			\end{minipage}
		}&%
		\subfigure[BM3D ($\sigma=20$) + SRN]{
			\begin{minipage}[!t]{0.2\linewidth}
				\centering
				\includegraphics[width=1.2in,height=0.6in]{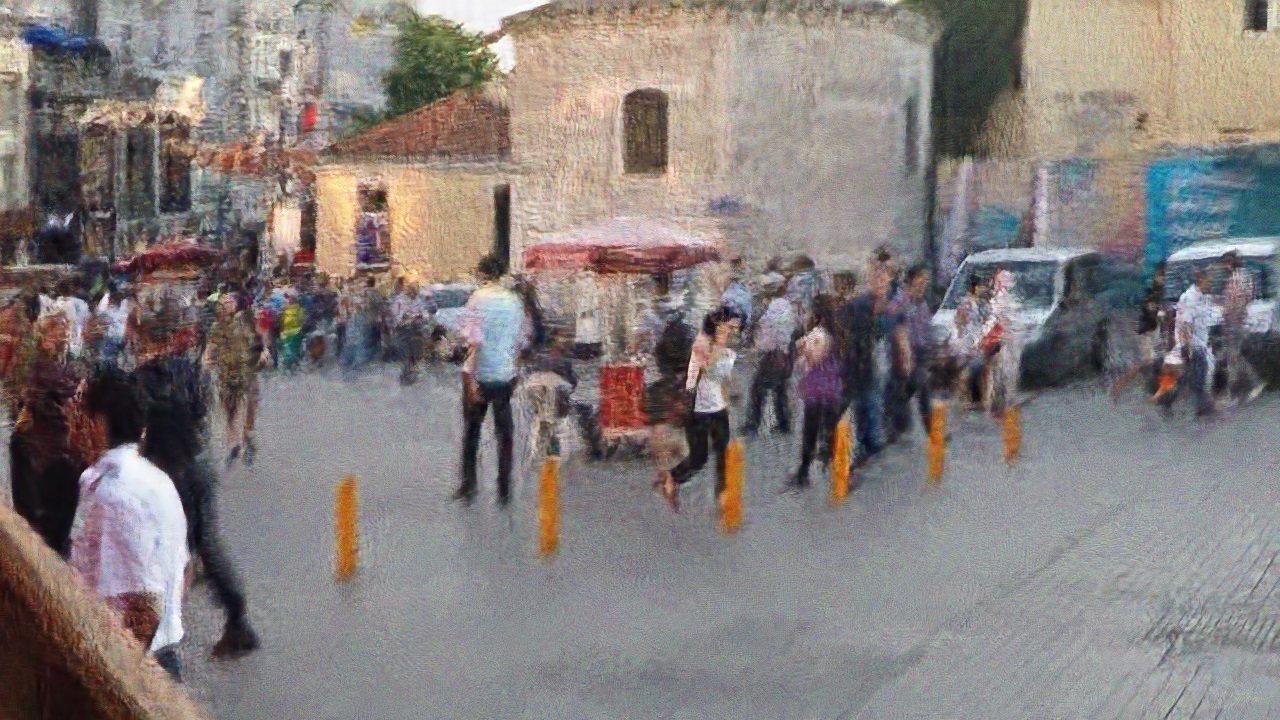}
			\end{minipage}
		}\\%
		\subfigure[BM3D ($\sigma=30$)]{
			\begin{minipage}[!t]{0.2\linewidth}
				\centering
				\includegraphics[width=1.2in,height=0.6in]{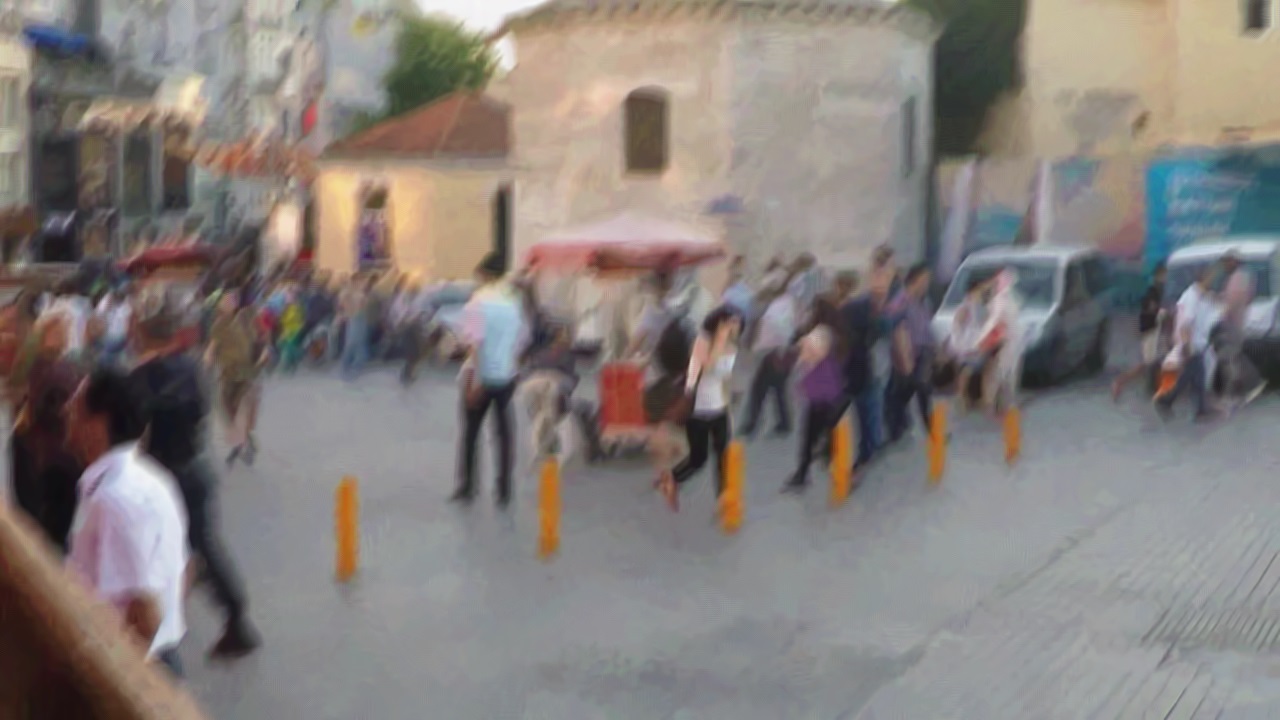}
			\end{minipage}
		}&%
		
		\subfigure[BM3D ($\sigma=30$) + SRN]{
			\begin{minipage}[!t]{0.2\linewidth}
				\centering
				\includegraphics[width=1.2in,height=0.6in]{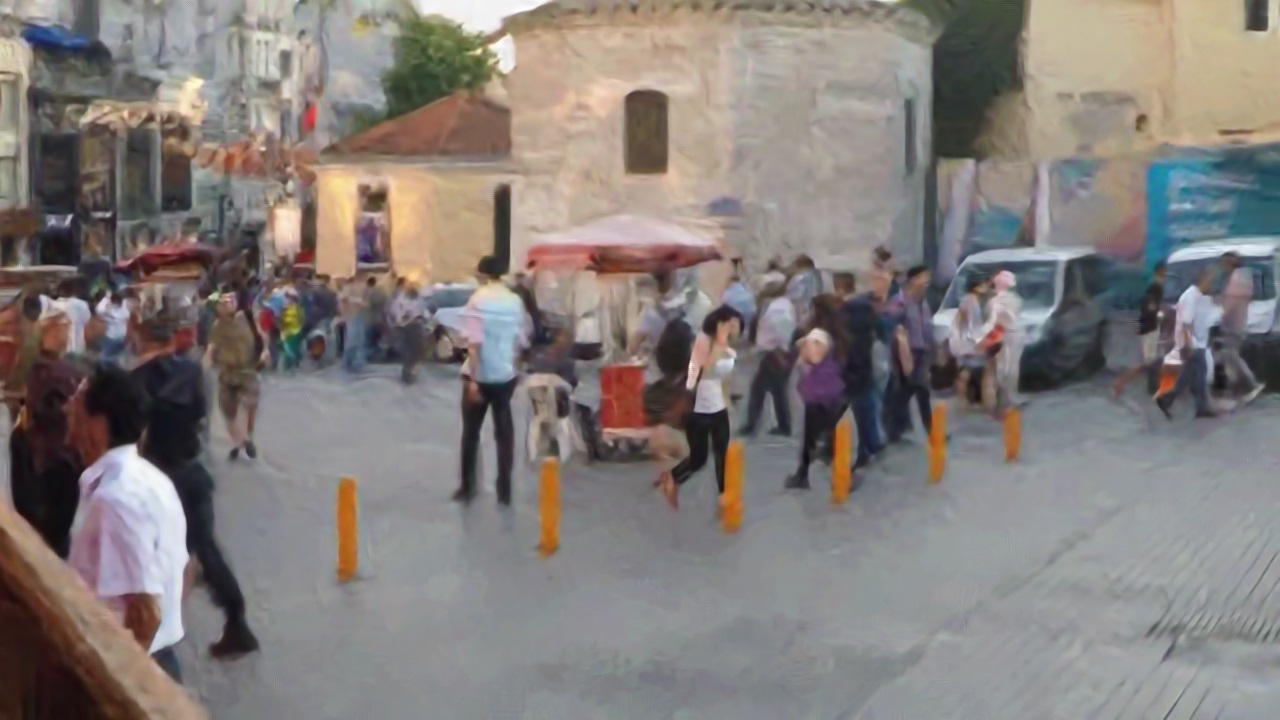}
			\end{minipage}
		}&%
		\subfigure[Ours]{
			\begin{minipage}[!t]{0.2\linewidth}
				\centering
				\includegraphics[width=1.2in,height=0.6in]{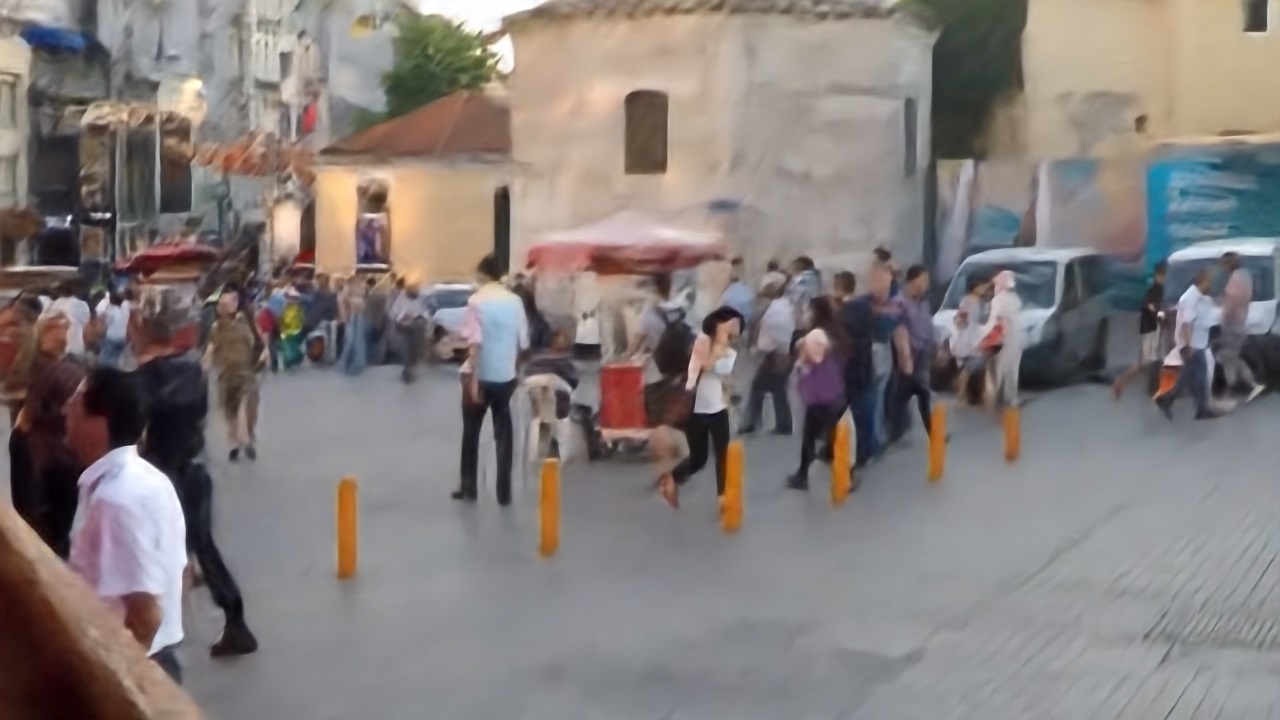}
			\end{minipage}
		}&%
		\subfigure[Ground truth ]{
			\begin{minipage}[!t]{0.2\linewidth}
				\centering
				\includegraphics[width=1.2in,height=0.6in]{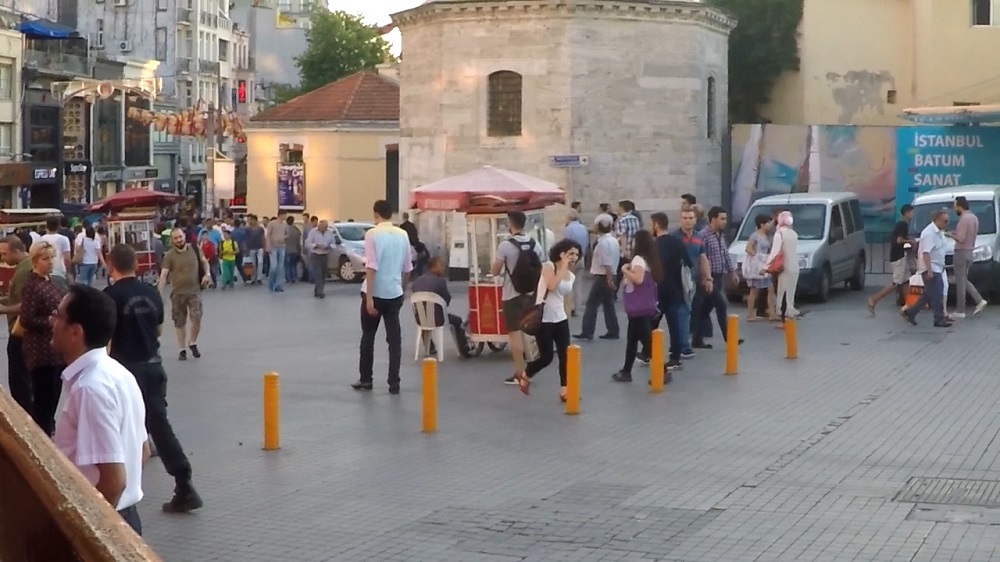}
			\end{minipage}
		}
	\end{tabular}
	\caption{(a) Input image. (b) SRN is unable to deblur if the image is noisy. (c, d) If the denoiser (BM3D) underestimates the noise level, the denoised image would remain noisy and these noise cause significant ringing artifacts in the deblurring stage. (e, f) Even if BM3D correctly estimates the noise level ($\sigma=30$), the residual noise still cause ringing artifacts in the deblurring stage. (g) Ours (h) Ground truth. BM3D+SRN means first denoising using BM3D then deblurring using SRN.\protect\footref{plus}}
	\label{noise}
\end{figure*}

\section{Proposed Method}
\label{sec:format}
\begin{figure*}[!t]
	\begin{center}
		\centerline{\includegraphics[width=6in,height=1.7in]{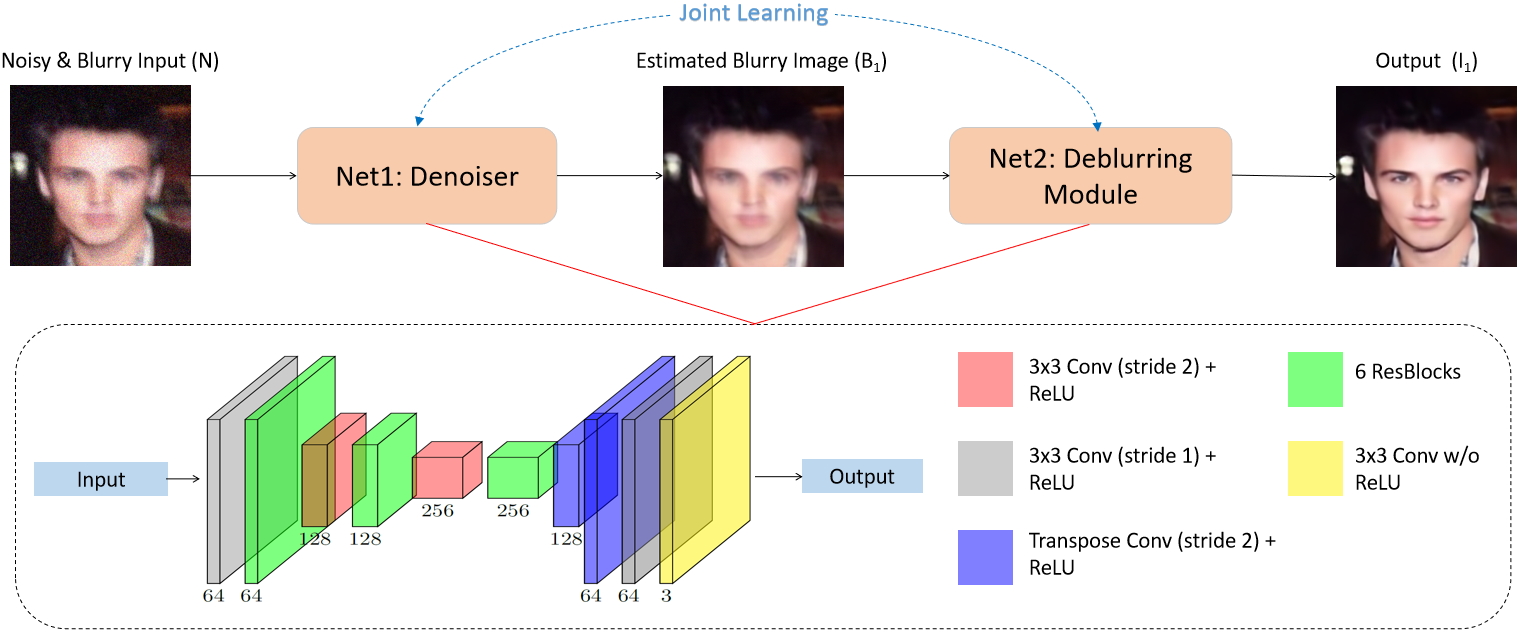}}
	\end{center}
    \vspace{-2em}
	\caption{An overview of the proposed NIDF. The two subnetworks use the same architecture but have different weights. }
	\label{architecture}
\end{figure*}

\subsection{Network Architecture}{ We present the NIDF architecture in Fig. \ref{architecture}\footref{link}. We discover that deblurring networks do not work well under noise, while denoisers are usually immune to blurs. Therefore, we concatenate the denoiser subnetwork (Net1) before the deblurring subnetwork (Net2). The proposed cascaded structure has two major advantages. First, its relatively light structure speeds up inference. Some approaches boost performance by exploiting sophisticated architectures, e.g., SRN \cite{SRN} employs a multi-scale structure, and \cite{Inception_GAN} uses multiple dense IRD blocks. These architectures benefit deblurring but do not help to handle noise. They also slow down inference. Second, compared with merging the two subnetworks into an all-in-one architecture, the cascaded structure allows Net1 and Net2 to work independently. Inspired by \cite{DeblurGAN, GFN, ED-DSRN}, we use a U-shape encoder-decoder \cite{UNet} that has been shown effective in image restoration for both subnetworks. Different from DeblurGAN \cite{DeblurGAN}, we do not employ adversarial training which is unstable. Compared to \cite{GFN, ED-DSRN} that use parallel branches for joint deblurring and super-resolution, we use the cascaded structure because the deblurring subnetwork performs better on denoised images than the original noisy images.
}

\subsection{Loss Functions}{
The proposed NIDF takes a noisy image $N$ as input. NIDF produces a denoised image $B_1$ and a sharp image $I_1$ simultaneously (Eq. \ref{NIDF}):
{
	\setlength\abovedisplayskip{1pt plus 3pt minus 7pt}
	\setlength\belowdisplayskip{1pt plus 3pt minus 7pt}
	\begin{equation}
	B_1 = Net1(N);\,I_1 = Net2(B_1);\,B_1,\,I_1 = NIDF(N). \label{NIDF}
	\end{equation}
}
We denote $B$ and $I$ as the ground truth images of $B_1$ and $I_1$. For pretraining, we define the loss function $\mathcal{L}_{denoiser}$ (Eq. \ref{denoiser_loss}) that is only related to the denoiser (Net1) in Fig. \ref{architecture}.
{
\setlength\abovedisplayskip{1pt plus 3pt minus 7pt}
\setlength\belowdisplayskip{1pt plus 3pt minus 7pt}
\begin{equation}
\mathcal{L}_{denoiser}(B, B_1) = ||B-B_1||_2^2. \label{denoiser_loss}
\end{equation}
}
Similarly, we define the loss function $\mathcal{L}_{deblurring}$ (Eq. \ref{deblurring_loss}) that is only related to Net2.
{
\setlength\abovedisplayskip{1pt plus 3pt minus 7pt}
\setlength\belowdisplayskip{1pt plus 3pt minus 7pt}
\begin{equation}
\mathcal{L}_{deblurring}(I, B) = ||I-Net2(B)||_2^2. \label{deblurring_loss}
\end{equation}
}
For joint learning, we define the joint loss function $\mathcal{L}_{joint}$ (Eq. \ref{joint loss}). By minimizing $\mathcal{L}_{joint}$, $I_1$ would get closer to $I$ no matter $B_1$ contains noise or not. Without $\mathcal{L}_{joint}$, Net2 is independent from Net1 and unable to output a sharp $I_1$ from a noisy $B_1$.
{
\setlength\abovedisplayskip{1pt plus 3pt minus 7pt}
\setlength\belowdisplayskip{1pt plus 3pt minus 7pt}
\begin{equation}
\mathcal{L}_{joint} = ||B-B_1||_2^2+ 0.5||I-I_1||_2^2. \label{joint loss}
\end{equation}
}
}

\footnotetext[2]{In our paper, "Denoiser+Deblurring Method" means first denoising then deblurring, vice versa.\label{plus}}

\subsection{Datasets Setup} {We choose 113831 training samples and 100 test samples from CelebA \cite{CelebA} to build a synthesized dataset. For each sharp face $I$, we generate a square PSF $P$ of side length $(2l+1)$ ($3 \leq l \leq 24$) using the random walk algorithm \cite{DeblurGAN}. We first resize $I$ to $256 \times 256$, then convolve $I$ with $P$ to acquire the blurry face $B$ (Eq. \ref{prepare}):
{\setlength\abovedisplayskip{1pt plus 3pt minus 7pt}
\begin{equation}
	B = I*P. \label{prepare} 
\end{equation}
\vspace{-2em}
} 

GOPRO \cite{DeepDeblur} dataset includes 2103 training images and 1111 test images (1280 $\times$ 720). For each sharp image $I$, the blurry image $B$ is generated by averaging the nearby 100 frames of $I$. Most blurs in the GOPRO dataset are non-uniform.

For both datasets, given the blurry image $B$, we generate the noisy image $N$ by adding AWGN: $N = B+n;\,n \sim \mathcal{N}(0, \sigma^2)$. $\sigma$ is chosen from $\{10,\, 20,\, 30,\, 40\}$ of equal possibilities. $N$ is the input of NIDF and $B$, $I$ are the corresponding ground truth images of $B_1$, $I_1$ in Eqs.\ref{NIDF}-\ref{joint loss}.
}

\subsection{Training Details}{During training stage, we first pretrain Net1 and Net2 separately. After they converge, we use the joint loss function $\mathcal{L}_{joint}$ to train Net1 and Net2 simultaneously. 

We use PyTorch to implement NIDF. All experiments are performed on an NVIDIA Tesla M40 GPU and a Xeon E5-2680 v4@2.40GHZ CPU with 256G memory. We use original face images from CelebA and $256\times256$ patches randomly cropped from GOPRO for training. The batch size is set to 16 and the optimizer is Adam \cite{Adam}. 

In the pretraining stage, we set the learning rate to $10^{-4}$. We train both networks for 150 epochs on GOPRO and 3 epochs on CelebA because the training set of CelebA is much larger than that of GOPRO. We input $N$ and $B$ into Net1 and Net2 respectively, then minimize $\mathcal{L}_{denoising}$ (Eq. \ref{denoiser_loss}) and $\mathcal{L}_{deblurring}$ (Eq. \ref{deblurring_loss}) to train the two subnetworks separately.

In the joint learning stage, we only input $N$ into NIDF. We train another 150 epochs on GOPRO and 5 epochs on CelebA by minimizing $\mathcal{L}_{joint}$ (Eq. \ref{joint loss}) with learning rate $10^{-5}$. 
}

\subsection{PSF Estimation}{
Given only a blurry and noisy image $N$, could we estimate the PSF $P$ in Eq. \ref{prepare}? The estimated blurry face $B_1$ and sharp face $I_1$ can be produced from $N$ by NIDF. According to Eq. \ref{prepare} and the convolution theorem, the estimated $P$ (denoted as $\hat{P}$) could be calculated as:
{
\setlength\abovedisplayskip{1pt plus 3pt minus 7pt}
\begin{equation}
\hat{P} = FFT^{-1}\left({FFT(B_1)}/{FFT(I_1)}\right). \label{deconv}
\end{equation}
}
However, it is difficult to generate very accurate $I_1$ and $B_1$ when noise is severe, therefore Eq. \ref{deconv} may not be precise enough. Pan et al. \cite{face_exemplars} proposed to guide face deblurring with exemplars. For a blurry face $B_1$, they search in databases to find an exemplar $S_1$ whose edges are close to those of $B_1$. Nevertheless, their method requires lots of searching and the ideal exemplars may not exist in the databases. In our work, we directly use $I_1$ as an exemplar of $B_1$ because $I_1$ usually preserves a fine face structure. The optimization objective is rewritten from \cite{face_exemplars}:
\begin{equation}
\min\limits_{\hat{P},\,I_l} \underbrace{||\hat{P} * I_l - B_1||_2^2}_{\text{Data\,Term}} + 0.002\underbrace{||\nabla I_l||_0}_{\text{Sharp\,Edges}} + 0.001\underbrace{||\nabla I_l - \nabla I_1||_2^2}_{\text{Exemplar\,Term}}\label{exemplar}
\end{equation}
where $I_l$ is the latent sharp image and $\nabla$ denotes the gradient operator. We use the half-quadratic splitting technique in \cite{face_exemplars} to estimate $\hat{P}$. Different from \cite{face_exemplars}, we do not need to generate facial contour masks. We show an example of PSF estimation in Fig. \ref{kernel}.

For other concatenations of denoisers and deblurring methods, we use the estimated sharp image as the exemplar of the denoised image to estimate the PSF via Eq. \ref{exemplar}. 

\begin{figure}[!t]
	\begin{center}
		\centerline{\includegraphics[width=2.7in,height=1.3in]{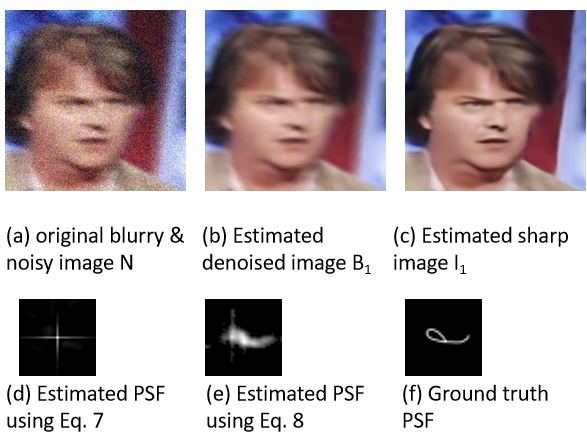}}
	\end{center}
	\vspace{-2em}
	\caption{An example of PSF estimation.}
	\label{kernel}
\end{figure}
}

\section{Experimental Results}
\label{sec:experimental}
We use Peak Signal-to-Noise Ratio (PSNR) and Structural Similarity Index (SSIM \cite{SSIM}) to evaluate NIDF. For the face deblurring task, we also use kernel similarity \cite{similarity} between the estimated PSF and the ground-truth PSF as another evaluation index. We evaluate NIDF on the synthesized CelebA dataset and the GOPRO dataset with additional noise. We compare to our method with three kinds of baselines: (1) Deblurring networks only; (2) Combinations of denoisers and deblurring networks; (3) NIDF without joint learning\footref{JL}, i.e., the two subnetworks in NIDF are trained separately. 

\begin{figure}[!t]
	\begin{center}
		\centering
		\subfigure[Input\newline (17.59/0.1323)]{
			\includegraphics[width=0.7in,height=0.6in]{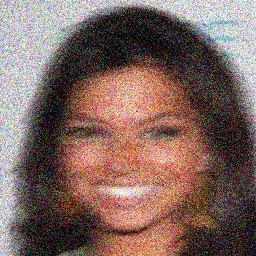}
		}
		\subfigure[\cite{FaceDeblur} + \cite{bm3d}\newline (23.39/0.6240)]{
			\includegraphics[width=0.7in,height=0.6in]{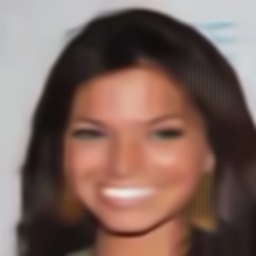}
		}
		\subfigure[\cite{CBDNet} + \cite{bm3d}\newline (18.53/0.5285)]{
			\includegraphics[width=0.7in,height=0.6in]{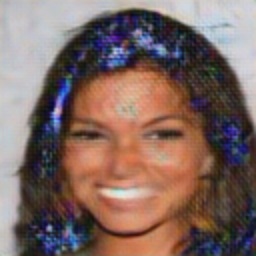}
		}
		\subfigure[\cite{bm3d} + \cite{FaceDeblur} \newline (23.60/0.6283)]{
			\includegraphics[width=0.7in,height=0.6in]{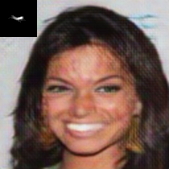}
		}\\%
		\subfigure[\cite{DnCNN} + \cite{FaceDeblur} \newline (21.25/0.5312)]{
			\includegraphics[width=0.7in,height=0.6in]{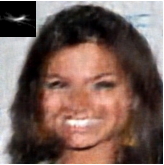}
		}
		\subfigure[w/o JL\footref{JL} \newline (22.60/0.5665)]{
			\includegraphics[width=0.7in,height=0.6in]{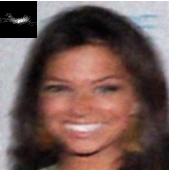}
		}
		\subfigure[NIDF \newline (24.40/0.6591)]{
			\includegraphics[width=0.7in,height=0.6in]{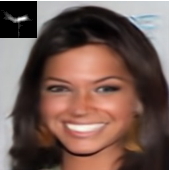}
		}
		\subfigure[Ground truth \newline (PSNR/SSIM)]{
			\includegraphics[width=0.7in,height=0.6in]{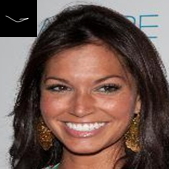}
		}
	\end{center}
	\vspace{-2em}
	\caption{Qualitative results on the CelebA dataset (input noise level $\sigma$=40).}
	\label{qualitative_CelebA}
\end{figure}

\begin{figure}[!t]
	\begin{center}
		\centering
		\subfigure[Input\qquad\qquad\newline (20.34/0.2646)]{
			\includegraphics[width=1.0in,height=0.6in]{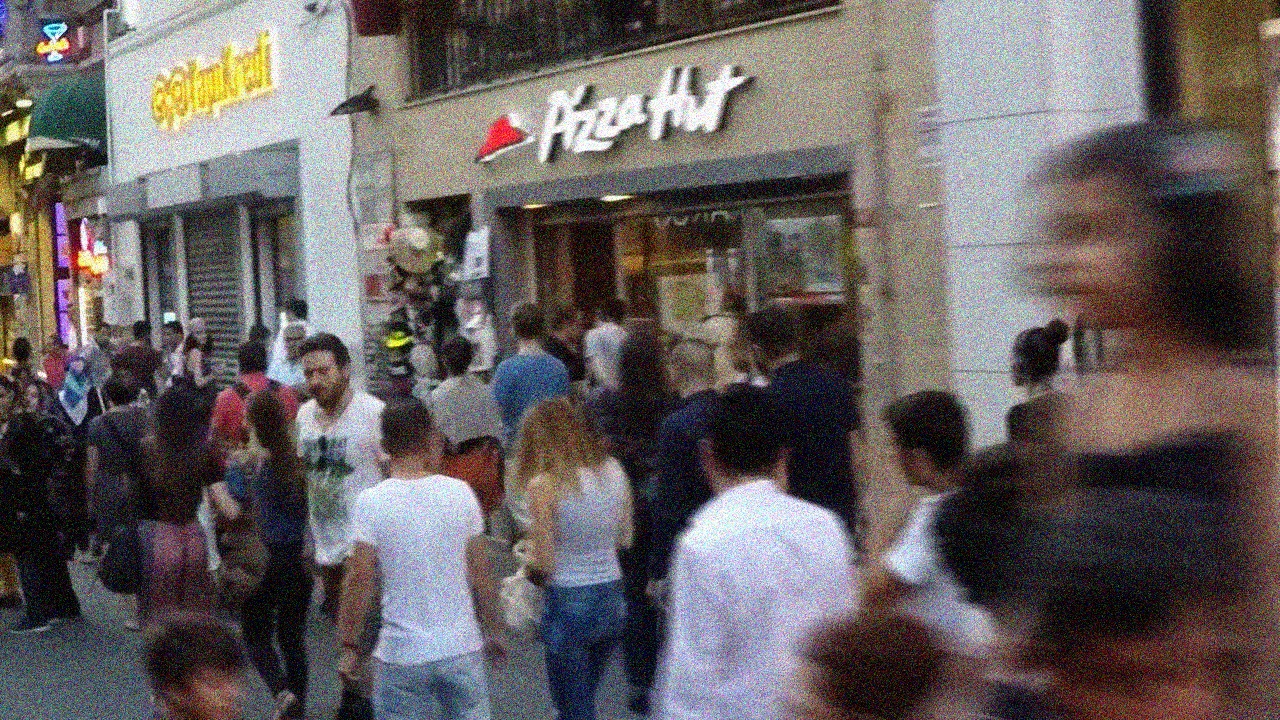}
		}
		\subfigure[\cite{bm3d}+\cite{SRN} \newline (24.69/0.8155)]{
			\includegraphics[width=1.0in,height=0.6in]{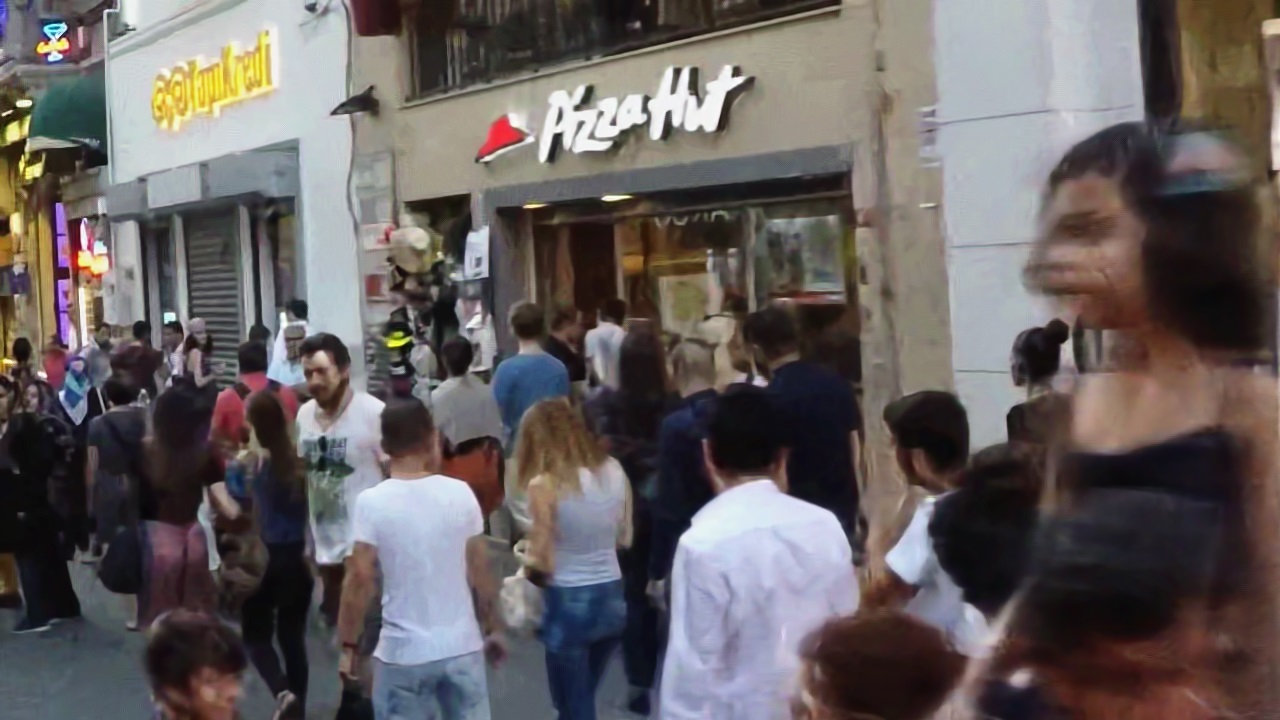}
		}
		\subfigure[\cite{DnCNN}+\cite{SRN} \newline (24.42/0.7801)]{
			\includegraphics[width=1.0in,height=0.6in]{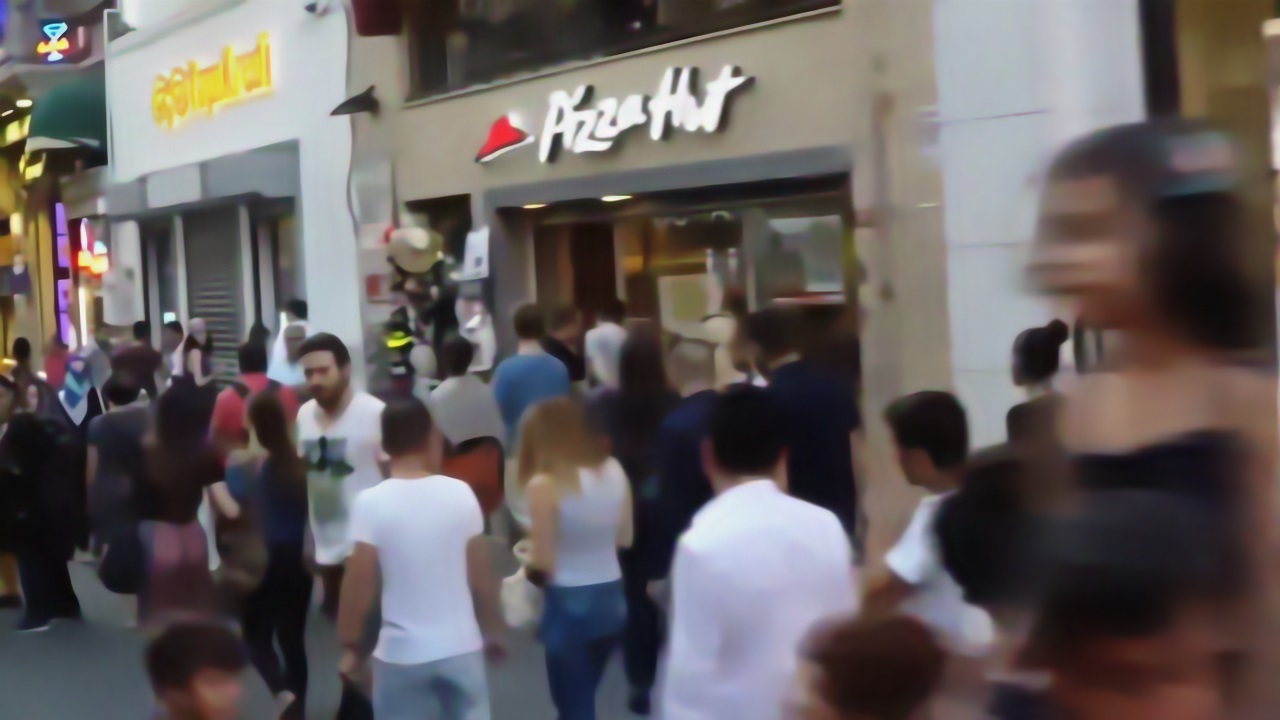}
		}\\%
		\subfigure[w/o JL\footref{JL}  \quad\newline (23.22/0.7248)]{
			\includegraphics[width=1.0in,height=0.6in]{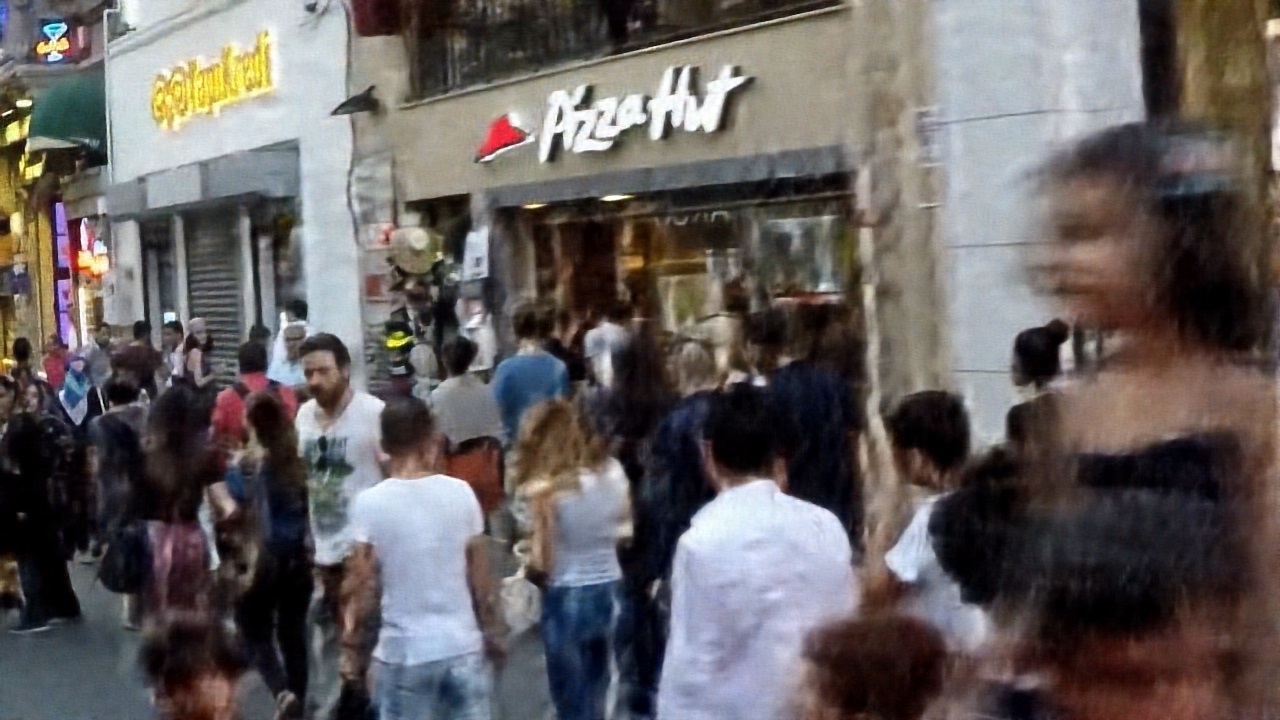}
		}
		\subfigure[NIDF \qquad\qquad \newline (25.40/0.8508)]{
			\includegraphics[width=1.0in,height=0.6in]{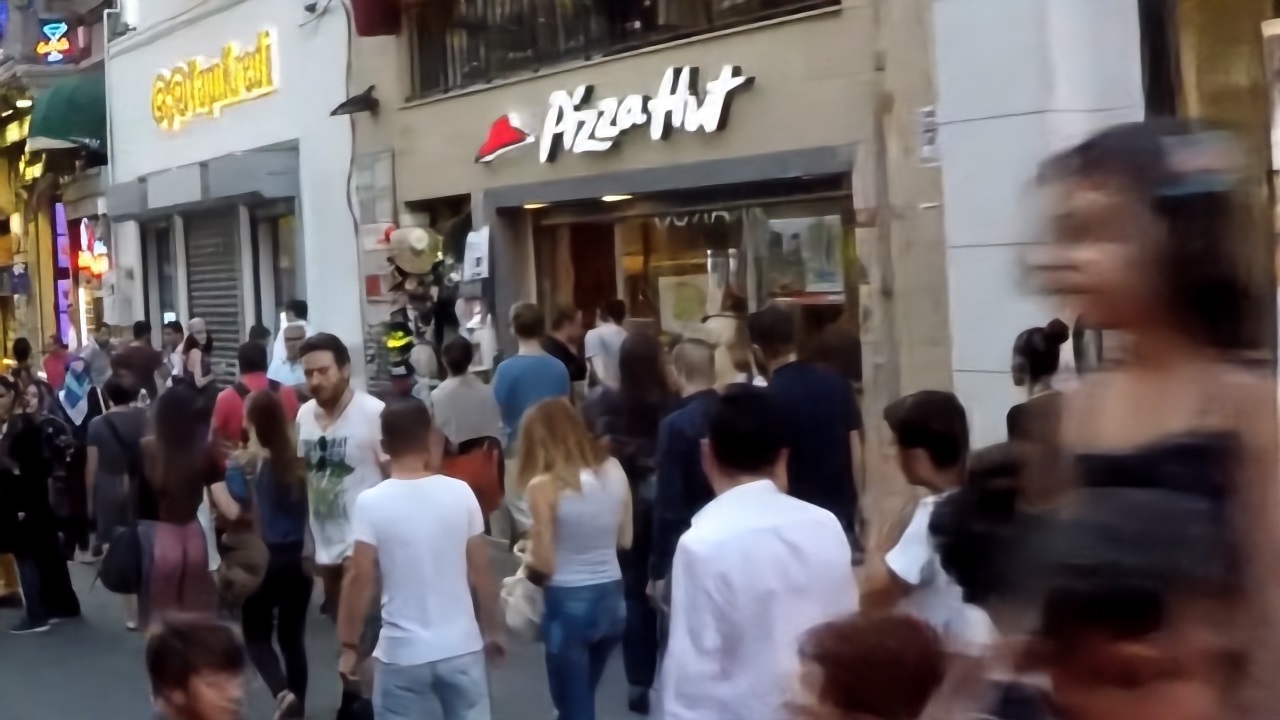}
		}
		\subfigure[Ground truth \newline (PSNR/SSIM)]{
			\includegraphics[width=1.0in,height=0.6in]{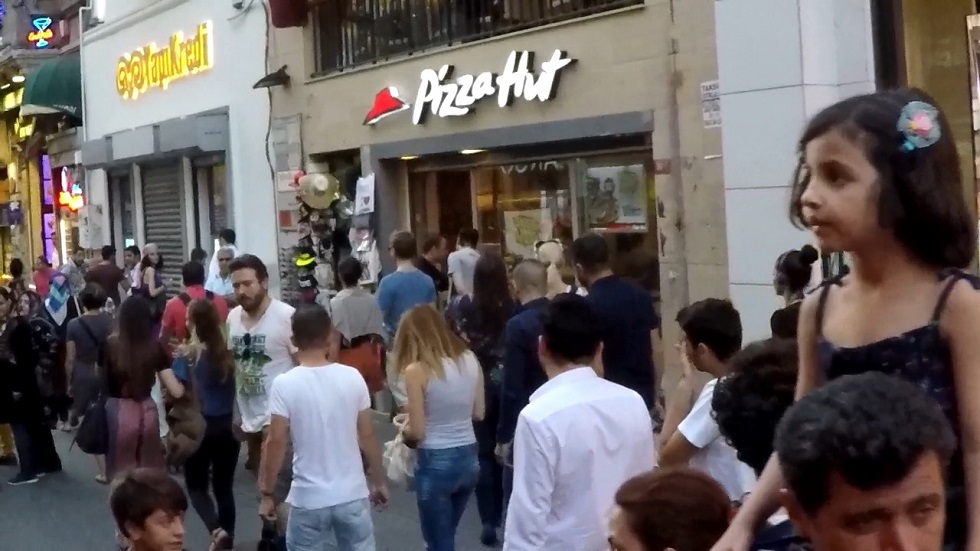}
		}
	\end{center}
	\vspace{-2em}
	\caption{Qualitative results on the GOPRO dataset (input noise level $\sigma$=30).}
	\label{qualitative_GOPRO}
\end{figure}

We report the quantitative and qualitative results on CelebA and GOPRO datasets in Tables \ref{quantitative_CelebA}-\ref{running_time} and Figs. \ref{qualitative_CelebA}, \ref{qualitative_GOPRO}. The concatenation methods, e.g., BM3D \cite{bm3d}+SRN \cite{SRN} and BM3D \cite{bm3d}+FaceDeblur \cite{FaceDeblur}, often introduce ringing artifacts because of the residual noise. Besides, we assume that noise levels are known in our experiments for fair comparisons. However, in real applications, degraded images are often contaminated with noise of unknown levels. When using BM3D or DnCNN, we have to estimate the noise levels very accurately to avoid artifacts, which is challenging and inconvenient. Conversely, we can directly use NIDF without image preprocess. Although BM3D \cite{bm3d}+SRN \cite{SRN} has higher PSNR than NIDF when noise are mild ($\sigma \leq 20$), NIDF is more effective under severe noise ($\sigma \geq 30$). Besides, BM3D+SRN is much slower than NIDF. We notice that NIDF (without joint learning) is inferior to NIDF (with joint learning) on both datasets. Therefore, the performance improvements of NIDF mainly owes to joint learning.

We have also tried to compare NIDF with CBDNet \cite{CBDNet} that is excellent at handling spatial-variant noise. However, CBDNet produces evident artifacts when noise and blurs are severe (Fig. \ref{qualitative_CelebA}(c)). Table \ref{similarity} shows that our proposed method is also beneficial to PSF estimation of the face deblurring task. 

\begin{table}[!t]
	\centering
	\caption{Average PSNR(dB)/SSIM on the CelebA dataset.\protect\footref{plus}}
	\label{quantitative_CelebA}
	\begin{tabular}{l|l|l|l|l|l}
		\multirow{2}{*}{Methods} & \multicolumn{4}{|c|}{PSNR vs noise level $\sigma$} & \multirow{2}{*}{SSIM} \\
		\cmidrule{2-5} &$\sigma$=10 & $\sigma$=20 & $\sigma$=30 & $\sigma$=40 \\
		\bottomrule[2pt]
		Input &23.97 &20.31 &18.03 &16.25 & 0.2654\\
		\hline
		\cite{FaceDeblur} &25.97 &24.13 &23.75 &22.61 & 0.6610\\
		\bottomrule[2pt]
		\hline
		\cite{FaceDeblur} + \cite{DnCNN} & 25.37 & 24.21 & 24.18 & 23.43 & 0.7004 \\
		\hline
		\cite{FaceDeblur} + \cite{bm3d} & 25.93 & 24.58 & 24.44 & 23.51 & 0.7242 \\
		\bottomrule[2pt]
		\hline
		\cite{DnCNN} + \cite{FaceDeblur} &25.94 &24.63 &24.36 &23.70 &0.7169 \\
		\hline
		\cite{bm3d} + \cite{FaceDeblur} &26.02 &24.13 &24.02 &23.11 &0.7092\\
		\bottomrule[2pt]
		w/o JL\footref{JL} &27.31 &25.17 &25.10 & 24.57&0.7432\\
		\hline
		NIDF &\textbf{28.64}  &\textbf{26.97} &\textbf{26.56} &\textbf{25.96} &\textbf{0.8105}\\
	\end{tabular}
\end{table}

\begin{table}[!t]
	\centering
	\caption{Average PSNR(dB)/SSIM on the GOPRO dataset.\protect\footref{plus}}
	\label{quantitative_GOPRO}
	\begin{tabular}{l|l|l|l|l|l}
		\multirow{2}{*}{Methods} & \multicolumn{4}{|c|}{PSNR vs noise level $\sigma$} & \multirow{2}{*}{SSIM} \\
		\cmidrule{2-5} &$\sigma$=10 & $\sigma$=20 & $\sigma$=30 & $\sigma$=40 \\
		\bottomrule[2pt]
		Input & 23.38 &20.34 &17.91 &15.96 & 0.2606\\ 
		\hline
		\cite{DeblurGAN} &23.19 &20.23 &17.86 &15.92 & 0.2545\\
		\hline
		\cite{SRN}  &24.75 &21.92 &20.11 & 18.53 & 0.3493\\
		\bottomrule[2pt]
		\cite{DeblurGAN} + \cite{DnCNN} &24.62 &24.54 &23.72 &23.64 & 0.6979 \\
		\hline
		\cite{SRN} + \cite{DnCNN} &26.47 &25.02 &24.59 &23.58 & 0.7146 \\
		\hline
		\cite{DeblurGAN} + \cite{bm3d} &25.13 &24.79 &24.63 &24.27 & 0.7267\\
		\hline
		\cite{SRN} + \cite{bm3d} &26.24 &25.13 &24.75 &24.30 & 0.7345 \\
		\bottomrule[2pt]
		\cite{DnCNN} + \cite{DeblurGAN} &25.39 &25.24  &23.23 & 23.07 & 0.7103\\
		\hline
		\cite{DnCNN} + \cite{SRN} &26.07 &25.95 &24.33 & 24.16 & 0.7222\\
		\hline
		\cite{bm3d} + \cite{DeblurGAN} &25.91 &25.55 &25.27 &24.70 & 0.7339\\
		\hline
		\cite{bm3d} + \cite{SRN}  &\textbf{27.02}&\textbf{26.43} &25.96 &25.32 & 0.7646\\
		\bottomrule[2pt]
		w/o JL\footref{JL} &25.96 &24.37 &23.44 &22.97 & 0.6723\\
		\hline
		NIDF  & 26.79 &26.27&\textbf{25.98}& \textbf{25.50} & \textbf{0.7759}\\
	\end{tabular}
\end{table}

\begin{table}[!t]
	\centering
	\caption{Average kernel similarity (KS)\cite{similarity} on the CelebA dataset.\footref{plus}}	
	\label{similarity}
	\begin{tabular}{l|l|l|l|l}
		& \cite{DnCNN} + \cite{FaceDeblur} & \cite{bm3d} + \cite{FaceDeblur} & w/o JL\footref{JL} & NIDF \\
		\hline
		KS & 0.6315 & 0.6348 & 0.6071 & \textbf{0.6498}
	\end{tabular}
\end{table}

\begin{table}[!t]
	\small
	\centering
	\caption{Running time (per image) on a Tesla M40 GPU.}
	\label{running_time}
	\begin{tabular}{l|l|l|l}
		Methods & Implementation &Second(s) & Resolution \\
		\hline
		DeblurGAN \cite{DeblurGAN} & PyTorch &0.38 &1280 $\times$ 720 \\
		\hline
		SRN \cite{SRN} & Tensorflow &6.51 &1280 $\times$ 720 \\
		\hline
		FaceDeblur \cite{FaceDeblur} & PyTorch&0.03 &256 $\times$ 256 \\
		\hline
		DnCNN \cite{DnCNN} & PyTorch & 0.60 & 1280 $\times$ 720 \\
		\hline
		DnCNN \cite{DnCNN} & PyTorch &0.29 & 256 $\times$ 256 \\
		\hline
		BM3D \cite{bm3d} & CUDA &0.54 &1280 $\times$ 720 \\
		\hline
		BM3D \cite{bm3d} & CUDA &0.21 &256 $\times$ 256 \\
		\hline
		NIDF & PyTorch&0.42 &1280 $\times$ 720 \\
		\hline
		NIDF & PyTorch &0.15 &256 $\times$ 256
	\end{tabular}
\end{table}

\section{Conclusion}
\label{sec:conclusion}
We propose a framework named NIDF to handle noise in image deblurring. Compared to previous deblurring methods, NIDF could tackle a more realistic deblurring problem where the blurred images contain noise. Our work has made three major contributions. First, joint learning of the two subnetworks in NIDF significantly improves the performance without increasing the model complexity. Second, NIDF does not require noise level estimation. Additionally, for the face deblurring task, NIDF could estimate the PSFs satisfactorily without searching exemplars. Third, extensive experiments show that our method is effective in terms of PSNR, SSIM, and kernel similarity \cite{similarity}. We find that Pan et al. \cite{Examplars} performs excellently under mild noise and small noise on PSF estimation, however it is inaccurate when the image is severely distorted. Our future work includes imporvements on \cite{Examplars}.

\footnotetext[3]{"NIDF without joint learning" is abbreviated to "w/o JL".\label{JL}}
\bibliographystyle{IEEEbib}
\bibliography{miao}

\end{document}